\documentclass{styles/svproc}
\usepackage{url}
\usepackage{comment}

\usepackage{amsmath, amsfonts, latexsym}
\usepackage{numprint}
\npdecimalsign{.}
\usepackage{hyperref}
\usepackage[nameinlink]{cleveref}
\usepackage{graphicx}
\usepackage{float}
\usepackage{afterpage}
\usepackage{multirow}
\usepackage{needspace}
\usepackage[all]{nowidow}
\usepackage[sort, numbers]{natbib}
\usepackage{tabularx}
\usepackage{threeparttable}
\usepackage[ruled,vlined,linesnumbered]{algorithm2e}
\SetAlgoCaptionSeparator{.}
\crefname{equation}{}{}
\Crefname{equation}{Equation}{Equations}
\crefname{section}{Sect.}{Sects.}
\Crefname{section}{Section}{Sections}
\crefname{figure}{Fig.}{Figs.}
\Crefname{figure}{Figure}{Figures}
\crefname{table}{Table}{Tables}
\Crefname{table}{Table}{Tables}
\crefname{algocf}{Alg.}{Algs.}
\Crefname{algocf}{Algorithm}{Algorithms}

\DeclareMathOperator{\Fone}{F_1}
\DeclareMathOperator{\KB}{KB}

\begin{document}
\mainmatter
\title{Topical Classification of Food Safety Publications with a Knowledge Base}
\titlerunning{Topical Classification of Food Safety Publications}
\toctitle{Topical Classification of Food Safety Publications with a Knowledge Base}
\author{Piotr Sowiński\inst{1,2} \and Katarzyna Wasielewska-Michniewska\inst{2} \and Maria Ganzha\inst{1,2} \and Marcin Paprzycki\inst{2}}
\authorrunning{P. Sowiński et al.}
\tocauthor{Piotr Sowiński, Katarzyna Wasielewska-Michniewska, Maria Ganzha, and Marcin Paprzycki}
\institute{Warsaw University of Technology, Warsaw, Poland\\
\email{piotr.sowinski@mini.pw.edu.pl}, \email{maria.ganzha@pw.edu.pl}
\and
Systems Research Institute, Polish Academy of Sciences\\
Warsaw, Poland\\
\email{\{katarzyna.wasielewska,marcin.paprzycki\}@ibspan.waw.pl}
}

\maketitle

\begin{abstract}
The vast body of scientific publications presents an increasing challenge of finding those that are relevant to a given research question, and making informed decisions on their basis. This becomes extremely difficult without the use of automated tools. Here, one possible area for improvement is automatic classification of publication abstracts according to their topic. This work introduces a novel, knowledge base-oriented publication classifier. The proposed method focuses on achieving scalability and easy adaptability to other domains. Classification speed and accuracy are shown to be satisfactory, in the very demanding field of food safety. Further development and evaluation of the method is needed, as the proposed approach shows much potential.

\end{abstract}
\section{Introduction}

In modern science we are confronted with a fast-growing body of publications, where finding the ``relevant ones'' is becoming increasingly challenging. To address the problem, evidence-based decision making (EBDM)~\cite{baba2012toward} was proposed, where one of the forms of research used is the systematic review (SR), which uses scientifically rigorous methods to identify, select, assess, and summarize multiple works~\cite{morton2011finding}. Systematic reviews are increasingly used to summarize the state-of-the-art and to guide the direction of future research~\cite{aromataris2014constructing}. While SRs are very reliable, due to the required time and expertise, producing them quickly proved to be a challenge~\cite{tsafnat2014systematic}; e.g. a study is typically included in a systematic review 2.5--6.5 years after publication~\cite{jonnalagadda2015automating}. Hence, the need for automation. While some parts of an SR require creativity and expertise, others can be automated~\cite{tsafnat2014systematic}.

This work discusses the use of semantic technologies and natural language processing to aid automation of SR in the field of food safety. More specifically, the described system facilitates relevance screening by classifying articles according to their topic(s). Having an article described by its title, abstract, and, optionally, unstructured keywords, the aim is to assign to it topics drawn from a knowledge base (KB). Here, potential advantage of using a semantic KB is explored, because of the meaning and the context provided by the ontology~\cite{salatino2019cso}.

The proposed solution was evaluated in the food safety domain, which presents a unique challenge in automatic SR, as it involves a wide variety of topics from multiple fields of science necessitated by the integration of heterogeneous sources of information~\cite{boqvist2018food,uyttendaele2016challenges}. On the other hand, there are numerous ontologies available for biology, chemistry, and medicine domains~\cite{whetzel2011bioportal}, which can be incorporated to reason about food safety.

\section{Related works} \label{sec:related_works}

Food safety is a relatively unexplored domain in text classification-oriented research. However, documents in this domain draw terminology from numerous biomedical sciences. Thus, one may expect to encounter similar characteristics to those present in related areas. For diseases and drugs, issues with entity linking usually stem from the many ways in which a single entity can be represented in text~\cite{li2017cnn}. There are examples of similar morphology but different meaning: ``ADA-SCID (adenosine deaminase deficiency)'' and ``X-SCID (X-linked combined immunodeficiency diseases)''. There are also examples of entirely different morphology, but similar semantic meaning, e.g. ``kaplan plauchu fitch syndrome'' and ``acrocraniofacial dysostosis''. Moreover, the numerous obscure abbreviations can vary between publications, bringing importance to the context. Finally, very few labeled datasets are available, making supervised methods unfeasible~\cite{zheng2015entity}.

On the other hand, in biomedical sciences there are many well-structured ontologies that may be used in entity linking~\cite{zheng2015entity,arbabi2019identifying}. There is a relatively large body of research concerning entity linking for medical reports and publications~\cite{li2017cnn}. However, these works usually focus on linking text to thesauri or taxonomies, such as MeSH\footnote{\url{https://www.nlm.nih.gov/mesh/introduction.html}} that are less expressive than full ontologies. Moreover, these methods are often evaluated using benchmark datasets that provide researchers with training data for supervised models. Nonetheless, even very sophisticated methods, based on BERT, can struggle to correctly perceive context and resolve ambiguous entities. Especially hard are situations where the model is asked to identify terms that occur rarely within the training corpus~\cite{fraser2019extracting}. Finally, modern supervised approaches have very high hardware requirements.

Furthermore, there a a few solutions addressing the classification problem in the biomedical domain. Here, we outline most relevant to the presented work.

NCBO Annotator\footnote{\url{https://bioportal.bioontology.org/annotator}} is an ontology-based tool for entity linking in biomedicine. Being a part of the BioPortal\footnote{\url{https://bioportal.bioontology.org/}}, it supports all ontologies published there. The Annotator uses basic string matching to find entities with names similar to those occurring in the text. Results are further enhanced with concept expansion and ontology mappings~\cite{jonquet2009ncbo}. Here, the method is unsupervised and can handle hundreds of ontologies at the same time. However, the system lacks the ability to handle morphological variants and more complicated disambiguation cases.

Neural Concept Recognizer~\cite{arbabi2019identifying} links entities from medical ontologies. It is based on the idea that both text and ontologies can be embedded into a vector space and one can compute similarity metrics between vectors to find matching terms. Text is vectorized using word embeddings and a convolutional neural network. Concept embeddings are established with respect to the taxonomical structure of the ontology. Training is performed on an artificially made dataset, derived from labels present in the ontology. The approach allows training the classifier based only on the ontology and allows easy adaptation to a different domain. It is also claimed that the system can handle novel synonyms, contrary to rule- or dictionary-based solutions. However, the authors point to the inability of the model to perceive context and a relatively slow performance.

The CSO Classifier~\cite{salatino2019cso} is an automatic classification system for computer science publications. Although it does not focus on the biomedical sciences, its general idea seems applicable to other domains as well. It uses the Computer Science Ontology of research topics, with a taxonomical structure~\cite{salatino2018computer}. The system consists of three modules. The syntactic module preprocesses the text and searches for n-grams, with high Levenshtein similarity to terms in the ontology. This is to find topics that are explicitly mentioned in the document. The semantic module aims at retrieving topics that are semantically related, but not explicitly mentioned. Part of speech tagging and a simple grammar rule are used for extracting candidate text spans to be analyzed. Next, word embeddings (via word2vec) are used to find semantically similar terms in the ontology. The most relevant topics are selected by ranking them using a metric that considers the frequency with which they occur in the document and how ``diverse'' were their mentions. Finally, in postprocessing, the terms most frequently occurring in the corpus are discarded and the remaining list of topics is enhanced by including direct ancestors of the terms from the ontology. According to the authors, the CSO Classifier outperforms other topic detection methods in terms of recall, by also in including topics that are only implied in text but not explicitly stated.

\section{Dataset preparation} \label{sec:data_food}

Let us now describe datasets that were used in our work. For constructing a corpus of food safety publications, two popular sources were used: PubMed\footnote{\url{https://pubmed.ncbi.nlm.nih.gov/}} and the European Food Safety Authority (EFSA) Journal\footnote{\url{https://www.efsa.europa.eu/en/publications}}. For both sources, articles relevant to food safety were found and their metadata retrieved.

\subsection{PubMed}

To retrieve the metadata from PubMed, the Entrez API client, from the Biopython package, was used~\cite{cock2009biopython}. Articles in PubMed are described using MeSH, which through its hierarchical structure groups descriptors into categories. Our search was limited to articles that use the ``food safety'' MeSH descriptor, and any of its children; \numprint{77348} articles were found, of which \numprint{61941} had available abstracts. \Cref{tab:pubmedtags} shows how often descriptors from a given MeSH category were used to point to articles in the corpus. The analysis was performed using SPARQL queries on MeSH and article metadata, imported into a Blazegraph quad store\footnote{\url{https://blazegraph.com/}}.

\begin{table}[htpb]
    \centering
    \caption{Descriptor usage by category for PubMed articles\label{tab:pubmedtags}}
    \begin{tabular}{l l r}
        \hline
        \multicolumn{2}{c}{\textbf{Category}} & \multicolumn{1}{c}{\textbf{\# descriptors}} \\
        \hline
        \multicolumn{2}{l}{Anatomy [A]} & \textbf{\numprint{26050}} \\*
        \textit{including:} & Fluids and Secretions & \numprint{4790} \\*
        & Cells & \numprint{4691} \\*
        & Plant Structures & \numprint{3770} \\*
        & Digestive System & \numprint{2519} \\
        \hline
        \multicolumn{2}{l}{Organisms [B]} & \textbf{\numprint{119840}} \\
        \hline
        \multicolumn{2}{l}{Diseases [C]} & \textbf{\numprint{45388}} \\*
        \textit{including:} & Infections & \numprint{15672} \\*
        & Animal Diseases & \numprint{5046} \\*
        & Digestive System Disorders & \numprint{4205} \\
        \hline
        \multicolumn{2}{l}{Chemicals and Drugs [D]} & \textbf{\numprint{209783}} \\
        \hline
        \multicolumn{2}{l}{Analytical, Diagnostic and Therapeutic Techniques (...) [E]} & \textbf{\numprint{101766}} \\*
        \textit{including:} & Investigative Techniques & \numprint{81670} \\
        \hline
        \multicolumn{2}{l}{Psychiatry and Psychology [F]} & \textbf{\numprint{5513}} \\
        \hline
        \multicolumn{2}{l}{Phenomena and Processes [G]} & \textbf{\numprint{128724}} \\
        \hline
        \multicolumn{2}{l}{Disciplines and Occupations [H]} & \textbf{\numprint{12752}} \\
        \hline
        \multicolumn{2}{l}{Anthropology, Education, Sociology (...) [I]} & \textbf{\numprint{8712}} \\
        \hline
        \multicolumn{2}{l}{Technology, Industry, and Agriculture [J]} & \textbf{\numprint{69369}} \\*
        \textit{including:} & Food and Beverages & \numprint{35432} \\*
        & Technology, Industry, and Agriculture & \numprint{31536} \\
        \hline
        \multicolumn{2}{l}{Humanities [K]} & \textbf{\numprint{1079}} \\
        \hline
        \multicolumn{2}{l}{Information Science [L]} & \textbf{\numprint{6161}} \\
        \hline
        \multicolumn{2}{l}{Named Groups [M]} & \textbf{\numprint{9822}} \\
        \hline
        \multicolumn{2}{l}{Health Care [N]} & \textbf{\numprint{111808}} \\
        \hline
        \multicolumn{2}{l}{Geographicals [Z]} & \textbf{\numprint{30474}} \\
        \hline
    \end{tabular}
\end{table}

It can be observed that food science is \textit{not} the dominant area. Even in [A] and [C] categories, terms associated with the digestive system are not the most frequent. Among frequent terms, not related to food science, are: organisms, chemicals, investigative techniques, phenomena and processes, health care, and geo-location. Thus, it is obvious that \textit{food safety touches upon a very diverse set of topics and involves complex relations}. Hence, when choosing ontologies for describing this domain, a very wide spectrum of topics has to be included.

To identify additional keywords that were not provided explicitly with the publications, named entity linking (NEL) of abstracts, against the UMLS metathesaurus, was performed with the scispaCy~\cite{neumann2019scispacy} library. The used model was based on BERT architecture (\texttt{en\_core\_sci\_scibert}), and the linker was targeting a $\sim$785k entity subset of UMLS. Finally, entity occurrence frequency in abstracts was summarized. The method identified \numprint{48083} unique entities in \numprint{61941} abstracts. The results contain some obvious misclassifications -- for example \numprint{28253} abstracts are to refer to mental concentration (attention concentration), which is a very uncommon topic in food safety. These occurrences most likely refer to the concentration of substances, which is a frequent topic in food safety literature. Barring the misclassifications, many of the frequently identified keywords refer to investigative techniques, risk assessment, and statistics (e.g. \textit{levels}, \textit{concentration}, \textit{detection}, \textit{contamination}, \textit{association}, \textit{significance}, \textit{prevalence}). Most other entities are typically within the scope of MeSH (biology, chemistry, medicine).

\subsection{EFSA}

EFSA metadata is available via a free API\footnote{\url{https://openapi-portal.efsa.europa.eu/}}, which allows retrieving abstracts and associated keywords. As the primary focus of EFSA is food safety, all available articles were retrieved. Of \numprint{10088} publications, \numprint{6684} had available abstracts.

Provided keywords are not structured, which prohibits analysis that was performed using MeSH descriptors. On the other hand, these keywords are not restricted by a controlled vocabulary and thus a rudimentary frequency analysis may show often used terms that were not identified previously. Among the most common keywords that are not directly within the scope of food science, are: \textit{risk assessment}, \textit{safety}, \textit{health claims}, \textit{MRL} (maximum residue levels), \textit{QPS} (qualified presumption of safety), \textit{exposure}, \textit{quarantine}, \textit{children}, \textit{data collection}, \textit{confirmatory data}, European Union member states, and various EU legislative documents. The latter appeared in contributions dealing with regulation compliance. It should be noted that nearly 67\% of keywords appear only once, while 14\% appear twice. These rarely-appearing terms are mostly specific chemicals, processes, and organisms.

Entity linking against UMLS was performed for the retrieved abstracts, using the same method as with the PubMed corpus. For \numprint{6684} abstracts, \numprint{13200} unique entities were found. As in PubMed, there were many misclassified terms that are unlikely to appear frequently in the corpus. The used NEL method is based on simple n-gram vectorization of mentions and lacks the necessary context required to successfully disambiguate some terms. Additionally, it failed to identify terminology specific to the EFSA dataset (e.g. \textit{European Commission}, \textit{MRLS}, \textit{MON}), due to UMLS lacking these terms. On the other hand, typical biomedical terms were identified easily. Often found were references to food products, ingredients, organisms, age groups, chemicals, and diseases.

\subsection{Summary}

From the preliminary analysis it is obvious, that dictionaries (ontologies) used in food safety should span a wide scope of biomedical topics. Moreover, the preliminary NEL results obtained using scispaCy strongly suggest that disambiguating of named entities in texts may require the system to better perceive the context in which the term appears and how is it related to other terms. 

\section{Ontologies} \label{sec:ontologiesfoodsafety}

While initial results suggest that the use of semantic technologies could be beneficial to the task of publication classification, there is no public food safety ontology available. However, there are ontologies describing related fields, such as food science. One example is FoodOn -- a comprehensive food ontology~\cite{dooley2018foodon}, available through NCBO's BioPortal~\cite{whetzel2011bioportal}. FoodOn has numerous mappings to other ontologies in the BioPortal\footnote{\url{https://bioportal.bioontology.org/ontologies/FOODON/?p=mappings}}, which allows them to be used jointly. FoodOn is also a part of OBO Foundry\footnote{\url{http://obofoundry.org/}}, guaranteeing application of several quality-improving guidelines~\cite{smith2007obo}, especially important when using multiple ontologies at a time.

As shown previously, food safety publications touch upon many diverse topics within the biomedical domain. The number of available biomedical ontologies is also high -- NCBO's BioPortal claims to host 846 ontologies. Without the help of domain experts, the only viable method for ontology selection (and use) would have to be based on data-driven analysis of topics found in publications. Such analysis can be based on data gathered from PubMed, thanks to the presence of structured MeSH descriptors. Those descriptors can be mapped to related entities in other ontologies, using mappings from the BioPortal.

For each of the~\numprint{11693} MeSH descriptors, present in the corpus, mappings to all BioPortal ontologies were retrieved. The initial focus was on ontologies from the OBO Foundry. Actively maintained ontologies with comprehensive documentation were prioritized. Additionally, several upper level ontologies were included (e.g., BFO, IAO, RO). The selected OBO ontologies are listed below, ordered alphabetically by their abbreviations.

\begin{itemize}
    \item Anatomical Entity Ontology (\textbf{AEO})\footnote{\url{http://www.obofoundry.org/ontology/aeo.html}}
    \item Agronomy Ontology (\textbf{AGRO})\footnote{\url{https://github.com/AgriculturalSemantics/agro}}
    \item Apollo Structured Vocabulary (\textbf{APOLLO-SV})\footnote{\url{https://github.com/ApolloDev/apollo-sv}}
    \item Basic Formal Ontology (\textbf{BFO})\footnote{\url{http://basic-formal-ontology.org/}}
    \item BRENDA tissue / enzyme source (\textbf{BTO})\footnote{\url{http://www.obofoundry.org/ontology/bto.html}}
    \item Common Anatomy Reference Ontology (\textbf{CARO})\footnote{\url{https://github.com/obophenotype/caro/}}
    \item Chemical Entities of Biological Interest (\textbf{CHEBI})\footnote{\url{https://www.ebi.ac.uk/chebi/}}
    \item Chemical Methods Ontology (\textbf{CHMO})\footnote{\url{http://obofoundry.org/ontology/chmo.html}}
    \item Cell Ontology (\textbf{CL})\footnote{\url{https://obophenotype.github.io/cell-ontology/}}
    \item Human Disease Ontology (\textbf{DOID})\footnote{\url{https://disease-ontology.org/}}
    \item Drug Ontology (\textbf{DRON})\footnote{\url{https://github.com/ufbmi/dron}}
    \item Human developmental anatomy ontology (\textbf{EHDAA2})\footnote{\url{http://obofoundry.org/ontology/ehdaa2.html}}
    \item Environment Ontology (\textbf{ENVO})\footnote{\url{http://environmentontology.org/}}
    \item Food-Biomarker Ontology (\textbf{FOBI})\footnote{\url{http://www.obofoundry.org/ontology/fobi.html}}
    \item FoodOn\footnote{\url{https://foodon.org/}}
    \item Gazetteer (\textbf{GAZ})\footnote{\url{http://obofoundry.org/ontology/gaz.html}}
    \item Gene Ontology (\textbf{GO})\footnote{\url{http://geneontology.org/}}
    \item Human Phenotype Ontology (\textbf{HP})\footnote{\url{http://www.obofoundry.org/ontology/hp.html}}
    \item Information Artifact Ontology (\textbf{IAO})\footnote{\url{https://github.com/information-artifact-ontology/IAO/}}
    \item Mammalian Phenotype Ontology (\textbf{MP})\footnote{\url{http://www.informatics.jax.org/vocab/mp_ontology/}}
    \item NCBI organismal classification (\textbf{NCBITaxon})\footnote{\url{http://obofoundry.org/ontology/ncbitaxon.html}}
    \item Ontology for Biomedical Investigations (\textbf{OBI})\footnote{\url{http://obi-ontology.org/}}
    \item Phenotype And Trait Ontology (\textbf{PATO})\footnote{\url{http://obofoundry.org/ontology/pato.html}}
    \item Population and Community Ontology (\textbf{PCO})\footnote{\url{https://github.com/PopulationAndCommunityOntology/pco}}
    \item Plant Experimental Conditions Ontology (\textbf{PECO})\footnote{\url{http://www.obofoundry.org/ontology/peco.html}}
    \item Plant Ontology (\textbf{PO})\footnote{\url{http://www.obofoundry.org/ontology/po.html}}
    \item Relation Ontology (\textbf{RO})\footnote{\url{https://oborel.github.io/}}
    \item Symptom Ontology (\textbf{SYMP})\footnote{\url{http://symptomontologywiki.igs.umaryland.edu/}}
    \item Uberon\footnote{\url{http://uberon.github.io/}}
    \item Units of measurement ontology (\textbf{UO})\footnote{\url{https://github.com/bio-ontology-research-group/unit-ontology}}
    \item Experimental condition ontology (\textbf{XCO})\footnote{\url{https://rgd.mcw.edu/rgdweb/ontology/view.html?acc_id=XCO:0000000}}
\end{itemize}

This set of ontologies covers \numprint{4031} out of \numprint{11693} (34.5\%) MeSH descriptors present in the corpus (\numprint{271714} out of \numprint{872838} (31.1\%) descriptor occurrences). This relatively low coverage contrasts with the very wide range of topics these ontologies cover. This is possibly due to MeSH being designed for indexing publications, and thus having terms oriented toward more general topics instead of granular concepts. For example, one of the frequently occurring MeSH descriptors is \textit{Fishes}, an imprecise term that does not have a strict definition. The NCBITaxon ontology includes many species of fish (and their biological taxonomy), but not this particular term. Other OBO Foundry ontologies also describe specific terms very well, but lack the more general/imprecise ones.

One way to cope with this issue would be to include an additional, ``meta-level'' biomedical ontology. MeSH could play this role, but it has several significant drawbacks. First, being a thesaurus, it is less expressive than ontologies. Second, its non-standard structure hinders its reuse with OWL ontologies that adhere to different design principles. Ultimately, the Systematized Nomenclature of Medicine Clinical Terms (SNOMED~CT) was selected. It covers well the ``general'' biomedical terms and contains very expressive relations~\cite{donnelly2006snomed}.

To integrate it with OBO Foundry ontologies, the terminology was converted to OWL\footnote{\url{https://github.com/IHTSDO/snomed-owl-toolkit}} and inserted into a knowledge base. Next, connections between identical (or related) terms in OBO Foundry ontologies and SNOMED~CT were created, using two sources of mappings. Some OBO ontologies already contained cross-database references to SNOMED~CT -- these references were normalized and converted to \texttt{skos:closeMatch} relations. Additionally, all mappings for SNOMED~CT were retrieved from the BioPortal API. In total, \numprint{43595} unique mappings were inserted into the KB, covering \numprint{37999} of all \numprint{354318} entities (10.7\%) in SNOMED~CT. From here, the combined OBO Foundry with SNOMED~CT knowledge base will be referred to as \mbox{\textbf{OBO/SNOMED}}.

MeSH descriptor coverage was then re-evaluated. The new set of ontologies covers~\numprint{7696} out of \numprint{11693} (65.8\%) MeSH descriptors, which constitutes \numprint{569104} out of \numprint{872838} (65.2\%) descriptor occurrences. The twenty most frequently occurring, yet uncovered MeSH descriptors include: \textit{food contamination}, \textit{food microbiology}, \textit{food safety}, \textit{pesticide residue}, and other terms directly related to food safety. This implies that the constructed ontology is still incomplete in this regard. This problem is not solvable with the immediately available resources, and solving it is out of scope of this contribution.

The knowledge graph was inspected for inconsistencies and other issues that may hamper the classification algorithm. The first discovered issue was the lack of consistency among cross-ontology references. They are supposed to connect similar/identical terms originating from different ontologies. Some such references are specified with loose annotation properties pointing to textual identifiers, not their URIs. Some target terms did have their identifiers attached using another annotation property, but many did not. These problems were mitigated with SPARQL UPDATE queries that transformed OBO references into traversable \texttt{skos:closeMatch} properties. Note that these issues are known and their detailed discussion can be found in the work of Laadhar et al.~\cite{laadhar2020investigating}. Other issues, e.g. triples with properties pointing to erroneous URIs, object properties referring to literal values, and annotation properties referring to objects, were also fixed with SPARQL queries. Such errors would decrease the accuracy of the method, due to it relying on the ontology for domain knowledge. 

At the end, the OBO/SNOMED graph became relatively large, containing almost 49 million triples. Handling such a large dataset presents significant challenges. Therefore, the graph was inspected for triples of little value to text-based information retrieval. Unnecessary information was removed including provenance information, metadata, links to external databases, and more. This resulted in a graph with approximately 25 million triples.

\section{Proposed approach} \label{c:method}

Let us now describe the method used to perform topical publication classification. Proposed approach is based on the following assumptions. (1) Classification should be as accurate as possible. (2) Solution should be fast and scalable, as target ontologies, depending on the discipline, can have tens of millions of triples. (3) The method should robustly cope with larger-than-memory datasets, without significant performance degradation. (4) Solution should be easily applicable to any domain. This implies reliance on the information contained in the KB, while ``domain-specific code'' should be limited to a minimum.

\begin{figure}[htb]
\includegraphics[width=10cm]{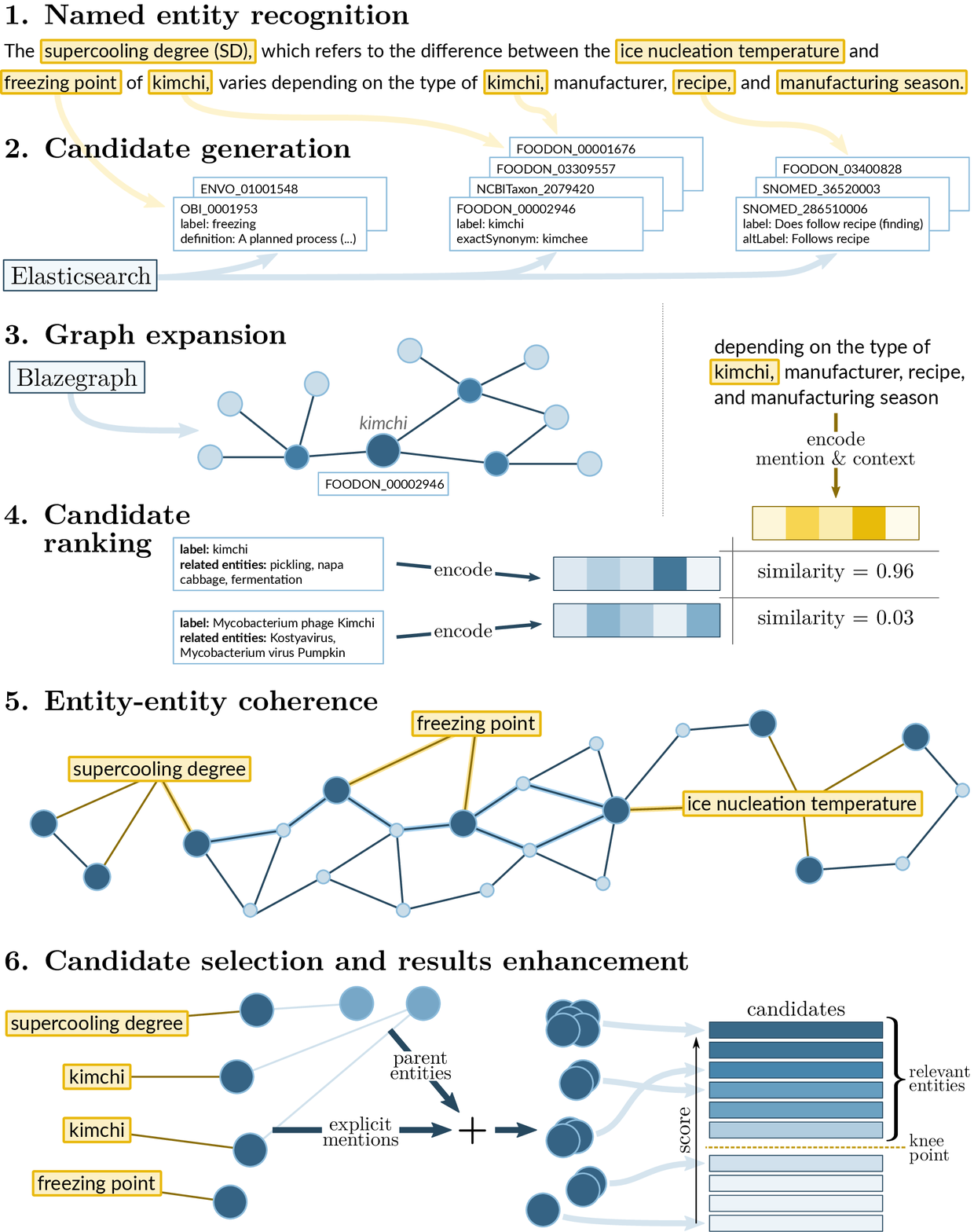}
\centering
\caption{Overview of the method}
\label{fig:method_diagram}
\end{figure}

\Cref{fig:method_diagram} presents an overview of the method. First, using named entity recognition (NER), mentions are identified in the text (\cref{sec:method_ner}). Then, for each mention, a set of candidate entities is produced, based on a full-text query to a search engine (\cref{subsec:cand_gen}). Neighborhoods of candidates are taken into account, using an algorithm that exploits the semantic meaning of relations between entities (\cref{subsec:graph_exp}). Candidates are compared to the mention, and each is assigned a similarity score (\cref{sec:cand_ranking}). Additionally, connections between candidates are used to discard the least coherent choices (\cref{sec:method_coherence}). Finally, the list of entities is enhanced using information from the KB, and the final set of terms most relevant to the publication is returned (\cref{subsec:cand_sel}). Let us now describe each operation in more detail.

\subsection{Named entity recognition} \label{sec:method_ner}

The first stage of the method identifies mentions, in publication abstracts that may correspond to ``entities of interest''. In later stages, these mentions will be linked to the entities in the knowledge base. There are several ready solutions to this problem in the biomedical domain, such as the scispaCy biomedical pipeline (\texttt{en\_core\_sci\_lg}), which achieves a 69\% $\Fone$ score in NER, on the very demanding MedMentions dataset~\cite{neumann2019scispacy} and it was chosen to be used here. The final set of mentions is additionally extended with explicitly provided keywords, which are available for the EFSA corpus.

\subsection{Candidate generation}\label{subsec:cand_gen}

Selecting candidate entities for a mention is not trivial when the number of entities in the KB is large; e.g. OBO/SNOMED is too large to be loaded into memory of a standard computer. Thus, Elasticsearch\footnote{\url{https://www.elastic.co/what-is/elasticsearch}} was used for candidate generation. Specifically, an Elasticsearch index was constructed, fields of which depended on types of text properties present within the KB. The unique identifier for each record in the index is simply the entity's URI. For each type of text property, a numeric weight was assigned for use when searching, to indicate how relevant the field is expected to be. Additionally, several fields are included for caching results of compute-intensive operations, described in detail in Sects.~\ref{subsec:graph_exp} and~\ref{subsec:cand_sel}. Finally, the constructed index for the OBO/SNOMED knowledge base consists of~\numprint{3739476} valid entity linking targets.

\subsection{Graph expansion}\label{subsec:graph_exp}

A candidate entity does not provide context information by itself -- it is necessary to look at related entities to obtain a better sense of what it describes. This ``graph expansion'' approach was used successfully in a textual entailment system~\cite{yadav2020medical}. Before attempting to design an algorithm for this problem, KB was examined and several observations were made.
\begin{itemize}
    \item[\textbf{O1}:] Traversing the graph only along the edges' original directionality may be too limiting. For example, with the \texttt{rdfs:subClassOf} relations this would result in only traversing the class hierarchy upward. Thus, the algorithm should traverse edges in both directions.
    \item[\textbf{O2}:] Processing large graphs with long-tail distributions of node degrees is challenging~\cite{gonzalez2012powergraph}. Encountering a vertex with a large number of neighbors forces the algorithm to visit all of them. Moreover, the relative amount of information a neighbor provides can vary considerably. Within large KBs one can find entities with numerous connections (e.g. many instances of one class), which makes the meaning of such an entity intuitively less ``concentrated''. Conversely, an entity with fewer connections is more likely to have a more concrete and relevant meaning.
    \item[\textbf{O3}:] Similarly, if an entity has many connections of type A, but very few of type B, the B-type edges are likely provide more ``focused'' information. For instance, the \textit{analgesic} class from CHEBI is connected to 65 substances that can be used as analgesics. When expanding from the \textit{analgesic} entity, they may introduce ``noise''. On the other hand, connections such as \textit{analgesic is a subclass of drug} do provide important contextual information.
    \item[\textbf{O4}:] Finally, different types of properties have different meanings and can be intuitively translated to connections of varying importance. For example, the \texttt{skos:closeMatch} relation implies greater similarity than \texttt{rdfs:subClassOf} or \texttt{obo:RO\_0002604} (is opposite of).
\end{itemize}

Based on these observations, the following formula for calculating edge weights ($\mathrm{w_{rel}}$) was proposed~\cref{eq:e_weight_f}. 

\begin{equation}
    \mathrm{l}(o) = \left|\left(s', p', o\right) \in \KB \cup \left(o, p', o'\right) \in \KB\right|
\end{equation}
\begin{equation}
    \mathrm{g}(s, p, d) = 
    \begin{cases}
        \left|\left(s, p, o'\right) \in \KB\right| & \text{if } d=\text{'spo'} \\
        \left|\left(s', p, s\right) \in \KB\right| & \text{if } d=\text{'ops'}
    \end{cases}
\end{equation}
\begin{equation}
    \label{eq:e_weight_f}
    \mathrm{w_{rel}}(s, p, o, d) = w_p \, \mathrm{l}(o)^{f_l} \, \mathrm{g}(s, p, d)^{f_g}
\end{equation}

Here $s$, $p$, $o$ (subject, predicate, object) is a triple from the KB that may be inverted to allow traversing the graph in either direction (\textbf{O1}). In such case, the $d$ parameter is set to $ops$ to indicate the inversion, otherwise it is $spo$. The final weight consists of three components:
\begin{itemize}
    \item $w_p$ -- the constant base weight for property $p$ (\textbf{O4}). The base weights have to be picked manually, based on the user's expertise, ontology documentation, and the use of the property in practice. The value typically ranges from 0.1 for very close, identity-like ties to 4, for the loose connections.
    \item $\mathrm{l}(o)$ -- the total number of links to and from the object of the triple (\textbf{O2}). By considering only the number of links of the object, the algorithm discourages visiting nodes with high cardinality, while leaving the node is not penalized. This is additionally scaled by a tunable parameter, $f_l$ ($0.5$ by default).
    \item $\mathrm{g}(s, p, d)$ -- the number of links from subject $s$, through predicate $p$, in direction $d$ (\textbf{O3}). This is also scaled by a parameter, $f_g$ ($0.5$ by default).
\end{itemize}

In the equations, $s'$, $p'$, $o'$ refer to (respectively) any subject, predicate, and object matching the given triple pattern. Expressions in the form of $(s, p, o)$ indicate triple patterns present in the~KB.

To compute $\mathrm{w_{rel}}$ effectively, the following approach is used. First, for each node $x$ in the KB, $\mathrm{l}(x)$ is calculated and the result inserted back into the graph. Next, a list of all $(s, p, o)$ triples in the KB is generated, along with their $\mathrm{l}(s)$ and $\mathrm{l}(o)$. The list is enhanced with inverted triples, to enable traversing edges in both directions. This results in a list that may be much larger than the system memory available, which necessitates careful processing of the data (in parts). To be able to perform grouping operations effectively, the list is sorted on disk and repartitioned by $s$, producing partitions that can be safely grouped in isolation. Then, it is grouped by $s$, $p$, and $d$, which corresponds to the $\mathrm{g}(s, p, d)$ function. Computing $\mathrm{w_{rel}}$ is straightforward now, and is done by iterating over all obtained groups. For each group only $c_{max}$ (4 by default) connections with lowest weights are kept. The partitioned data is processed in parallel using the Dask Python library~\cite{dask}.

The algorithm outputs a list of triples with weights that when inserted into the KB yield a weighted, traversable graph. A weighted graph can be more generally framed as an attribute graph, where any edge between two nodes (a triple) can be annotated with additional information. Blazegraph implements a solution to this called Reification Done Right (RDR), in which any triple can be either the object or the subject of another triple~\cite{hartig2014foundations}.

Perhaps the most obvious solution to finding neighboring entities in the knowledge graph would be to employ SPARQL queries, however, this would be hardly a typical use case for this language. A better fit would be a Gather-Apply-Scatter (GAS) algorithm, similar to the PowerGraph approach~\cite{gonzalez2012powergraph}. GAS allows one to write high-performance, parallel graph analytics algorithms. Blazegraph implements this approach\footnote{\url{https://github.com/blazegraph/database/wiki/RDF\_GAS\_API}} via a Java API. The graph expansion algorithm was implemented using this method and compiled as a Blazegraph plugin. The algorithm obtains edge weights of triples using RDR. The code is a modified version of the breadth-first search algorithm, taking into account maximum allowed traversal depth and maximum distance of found neighbors. The search is executed in parallel, across multiple threads.

As the neighborhood of an entity depends solely on the ontology, and not the particular search being made, it is possible to perform graph expansion for each possible candidate beforehand, to speed up classification. Then, the neighboring entities are saved to the Elasticsearch index, along with the distances to them (in the \texttt{related\_entities} and \texttt{related\_entities\_weights} fields). This removes the need to run and query Blazegraph during runtime, improving classification speed and reducing the overall memory footprint.

\subsection{Candidate ranking} \label{sec:cand_ranking}

Obtained candidate entities are then rated by their relevance to the mention. The algorithm should consider both the context in which the mention appears, as well as the neighborhood of the candidate entity within the KB. The proposed solution uses two types of text vector representations -- semantic and lexical. The semantic component employs word embeddings provided by the scispaCy library. While word embeddings have superior performance to lexical-based methods, one must also consider that out-of-vocabulary words will not produce a vector and thus cannot be compared. Here, for scientific texts, it is unreasonable to expect the embedding model to cover all possible words, some of which may be highly specialized and/or rarely occurring. Hence, the lexical component is introduced, based purely on character-level similarity between text fragments. To construct a lexical vector, the text fragment is split into 3-4 character n-grams. Next, the unique n-grams are grouped by their hashes and counted.

To compute the score of a candidate, its and all of its neighbors' text properties are vectorized. This produces two matrices, $L_{n \times m_l}$ and $S_{n \times m_s}$, for lexical and semantic vectors, respectively. Here $n$ is the number of all text properties from all entities that are taken into account. The number of columns depends on the length of the vectors. The matrices are L2-normalized row-wise, with the exception of zero-length vectors. Additionally, a vector of distances to related entities is retrieved ($w_e$), with values starting from zero (for the candidate entity itself). As described in~\cref{subsec:cand_gen}, text properties can have varying importance, which is considered here in the form of $w_p$, a vector of property weights. Finally, the similarity between the mention and other text fragments is calculated in bulk using~\cref{eq:similarity_v}.
\begin{equation}
    \label{eq:similarity_v}
    d = w_p \frac{w_{l} \, \mathrm{a}\left( L l\right) + w_{sm} \, \mathrm{a}\left( S s_m\right) + w_{sc} \, \mathrm{a}\left( S s_c\right)}{1 + w_e}
\end{equation}
\begin{equation}
    \label{eq:similarity_act}
    \mathrm{a}(x) = \left( 1 + \exp{\left( \alpha - \beta x \right)} \right)^{-2}
\end{equation}

Here $l$, $s_m$, $s_c$ refer to the lexical vector of the mention, its semantic vector, and the semantic vector of the sentence the mention is in (its context), respectively. It is assumed that these vectors are L2-normalized. Moreover, $w_l$, $w_{sm}$, $w_{sc}$ are tunable weights for the three components. Next, raw cosine similarity values are put through the activation function $a$, presented in~\cref{eq:similarity_act}. It is a case of the generalized logistic function. Its purpose is to discard low similarity scores and ``promote'' good matches. The $\alpha$ and $\beta$ parameters can be tuned to control its steepness and translation along the X axis. The function will return values from the $(0, 1)$ interval. \Cref{eq:similarity_v} produces an $n$-length vector of similarity scores between the mention and each of the considered text fragments. The score is accumulated by simply summing the elements of the vector $d$. The actual code for calculating this formula uses several performance optimizations, such as a lookup table for the activation function. Additionally, numerically-heavy routines were compiled using the Numba JIT compiler~\cite{lam2015numba}.

To further speed up classification, the feature vectors for text from the KB are precomputed during indexing. They are then stored and retrieved during classification, minimizing the amount of text processing at runtime. The cached vectors can take up large amounts of memory. For the OBO/SNOMED KB problems occurred even on a computer with $\sim$32\,GB of memory, and for other KBs required memory may increase further. This is solved by storing the vectors using the Lightning Memory-Mapped Database (LMDB)\footnote{\url{https://symas.com/lmdb/}}, which offers very high random-read performance and supports concurrent reads. This approach, however, necessitates the use of an SSD as the backing storage medium, as traditional hard drives are unsuitable for rapid random reads.

As a result of this step, for each mention, an ordered list of candidates with assigned scores is obtained. The greater the score, the more likely the candidate is to be a good match for the mention.

\subsection{Entity-entity coherence}\label{sec:method_coherence}

To help disambiguate named entities, the entity-entity coherence algorithm, introduced by Hoffart et al.~\cite{hoffart2011robust} is used, albeit with several modifications -- most notably the preprocessing stage is different. Original entity-entity coherence values are based on Wikipedia article similarity, which produces a dense similarity matrix. This would be hard to implement efficiently here, given the high complexity of algorithms computing all-pairs distance matrices in graphs.

Therefore, first, the highest scoring candidate entities across all mentions are selected, to discard noisy low-confidence scores. A sparse similarity matrix is constructed, which represents how closely candidates are related to each other. This is established by effectively approximating the shortest path between each pair of the candidates, using their cached neighborhoods from the graph expansion step. Resulting matrix will not be very dense. However, the most important connections should be present. Finally, the set of candidates is narrowed down by greedily removing the least strongly connected candidates. The algorithm outputs a mapping of candidate entities to boost values, scaled to be greater or equal to 1. This boost is used to multiply the scores of those entities, possibly changing the ordering of candidates for each mention. The candidate ranking lists are then sorted again, to account for this change.

To enhance the performance of the method, several additional ``bookkeeping'' data structures and optimizations are introduced. Additionally, the last stage of the algorithm, being heavy in numerical computations, is compiled using the Numba JIT compiler.

\subsection{Candidate selection and results enhancement} \label{subsec:cand_sel}

Having identified and ranked candidate entities for each mention, the final step is to determine the most relevant terms that describe the publication. The classification result should also include concepts that are not explicitly mentioned in the text. This is achieved by using a technique similar to the one employed in the CSO Classifier~\cite{salatino2019cso}, where the final set of terms is enhanced with ``parent'' entities, higher up in the ``semantic hierarchy''. Depending on the ontology this may translate to, for example, a subclassing relationship or the \texttt{skos:broader} property. The list of parent entities is established using a SPARQL query with parameters, which allow for specifying parent properties or inverse parent properties, to accommodate both approaches to expressing such relationships. Additionally, each parent is given a weight $w_p = links^{-\alpha}, \text{ where } \alpha \in (0, 1)$. Here $links$ is the total number of links the parent has and $\alpha$ is a scaling factor, set to 0.3 by default. These parent entities along with their weights are saved to Elasticsearch during indexing, in the \texttt{parent\_entities} and \texttt{parent\_entities\_weights} fields. This information is later retrieved during classification.

Similarly to the CSO Classifier, the ranking algorithm aggregates the top-scoring candidates across mentions, increasing their score based on how frequently they were identified in text and with what diversity (number of unique mention lemmas). From the final list of candidates only a portion is kept using, which is by finding the ``knee point'' in the distribution of scores. The knee is established using the Kneedle algorithm, which can adjust to different situations robustly~\cite{Satopaa2011FindingA} and is shown to perform significantly better than any fixed cutoff value.

\subsection{Summary}\label{sec:method_sum}

The presented method makes extensive use of preprocessing to offload most of the computationally intensive tasks to run-only-once jobs and in turn, speed-up classification. Part of the data is cached in Elasticsearch, which is accessed over an HTTP API, with only two batch requests per classified document. The encoded text vectors' significant size would make transporting them over HTTP inefficient. Thus, they are stored in an embedded database (LMDB), which achieves zero-copy reads, owing to its memory-mapped architecture.

Adaptability to other domains is achieved through the extensive use of configuration settings and modular design. One can easily create new mention detectors, semantic vector encoders, and lexical vector encoders, with a few lines of Python code. Other aspects of the classifier, such as the various tunable parameters of algorithms presented above, can be changed using a YAML configuration file. The process of retooling the entire pipeline to an entirely new domain is largely reduced to changing the configuration file and providing an appropriate ontology.

\section{Experimental results} \label{sec:experiments}

A thorough quantitative classification accuracy evaluation of the classifier could not be performed, due to the lack of a labeled dataset. Thus, the scope of the assessment was limited to identifying obvious misclassifications that would negatively impact the precision metric. For this purpose, articles from the EFSA and PubMed datasets were classified and selected results examined manually. \Cref{tab:pubmed_example}~presents example output of the method, with two identified errors highlighted in bold.

\begin{table}[htbp]
    \centering
    \small
    \begin{flushleft}
    \begin{threeparttable}
    \caption{Example classification of a food safety PubMed article~\cite{Deutch2004DietaryCI}\label{tab:pubmed_example}}
    \begin{tabularx}{\textwidth}{ 
      l
      >{\raggedright\arraybackslash}X
      >{\raggedright\arraybackslash}X
      r
    }
        \hline
        \multicolumn{1}{c}{\textbf{URI}} & \multicolumn{1}{c}{\textbf{Label}} & \multicolumn{1}{c}{\textbf{Mention lemmas}} & \multicolumn{1}{c}{\textbf{Score}} \\\hline
        obo:CHEBI\_26208 & polyunsaturated fatty acid & blood fatty acid, plasma fatty acid, fatty acid profile & 82.88 \\\hline 
        obo:NCBITaxon\_29073 & Ursus maritimus & polar bear, polar bear intake & 43.93 \\\hline 
        obo:CHEBI\_35366 & fatty acid & fatty acid & 33.97 \\\hline 
        obo:ENVO\_02500036 & environmental pollution & pollution & 33.86 \\\hline 
        obo:GAZ\_00052766\tnote{1} & Illoqqortoormiut & East Greenland, Ittoqqortoormiit & 32.30 \\\hline 
        obo:CHEBI\_5692 & hexachlorobenzene & hexachlorobenzene & 31.69 \\\hline 
        obo:OBI\_0000181 & population & population & 29.20 \\\hline 
        obo:GAZ\_00001507 & Greenland & Greenland & 27.75 \\\hline 
        obo:NCBITaxon\_9709 & Phocidae & seal & 25.13 \\\hline 
        snomedct:19314006 & Seal (organism) & seal & 22.48 \\\hline 
        obo:FOODON\_03411343 & whale & whale & 22.45 \\\hline 
        \textbf{snomedct:733446001}\tnote{2} & \textbf{Canadian (ethnic group)} & \textbf{canadian} & 22.18 \\\hline 
        snomedct:226365003 & N-3 fatty acid (substance) & n-3 fatty acid & 19.81 \\\hline 
        obo:CHMO\_0002820 & concentration & concentration & 19.73 \\\hline 
        obo:CHEBI\_34852 & mirex & mirex & 17.69 \\\hline 
        obo:CHEBI\_61204 & docosapentaenoic acid & docosapentaenoic acid & 17.29 \\\hline
        obo:CHEBI\_34623 & chlordane & chlordane & 16.85 \\\hline
        obo:CHEBI\_27208 & unsaturated fatty acid & blood fatty acid, plasma fatty acid, fatty acid profile & 16.62 \\\hline
        \textbf{snomedct:223503004}\tnote{3} & \textbf{North America (geographic location)} & \textbf{North} & 16.27 \\\hline
        obo:PATO\_0000025 & composition & composition & 14.90 \\\hline
        obo:UBERON\_0001969 & blood plasma & plasma & 14.72 \\\hline
        obo:FOODON\_03412406 & bear & polar bear, polar bear intake & 14.49 \\\hline
        obo:GAZ\_00228284 & Northwest Atlantic Ocean coastal waters of Greenland & Greenland & 13.87 \\\hline
        \multicolumn{4}{c}{\textit{5 more results omitted}} \\\hline
    \end{tabularx}
    \begin{tablenotes}
        \item[1] Illoqqortoormiut is an alternative name of Ittoqqortoormiit. The classification is correct.
        \item[2] Misclassification: the text refers to the ``Canadian guideline levels'', not the ethnic group of Canadians.
        \item[3] Misclassification: the text refers to Northern Greenland (the North), not the North America. Although Greenland is in fact considered to be a part of North America, this ``technically correct'' entry is probably just the result of coincidence.
    \end{tablenotes}
    \end{threeparttable}
    \end{flushleft}
\end{table}

It can be observed that the method performs generally well, albeit it tends to frequently return terms that do not seem to have much importance, such as: \textit{species}, \textit{application}, \textit{food}, \textit{population}. Usually, it is able to perceive context, but can still fail in some cases when understanding of the entire sentence is needed for correctly identifying the mention. Unfortunately, this is hard to achieve with simple word embeddings. However, rarely occurring and very specific terms such as taxa, geographic locations, and chemicals are identified easily and seemingly with good accuracy.

The performance of the method was measured on a modern Linux workstation with a 3.6~GHz 6-core CPU, 32~GB of system memory and an NVMe SSD rated at \numprint{480000}~IOPS. From the EFSA and PubMed corpora 50 publications were selected randomly and classified in batch. Only the wall clock time taken by classification was measured, excluding the time required to load the code and the models into memory. The experiment was repeated ten times. Classifying the entire EFSA batch took \numprint{78.48}\,s in average, with the mean time to classify a publication at \numprint{1.57}\,s. For PubMed, classifying the batch took \numprint{80.49}\,s on average.

\begin{figure}[htb]
\includegraphics[width=10cm]{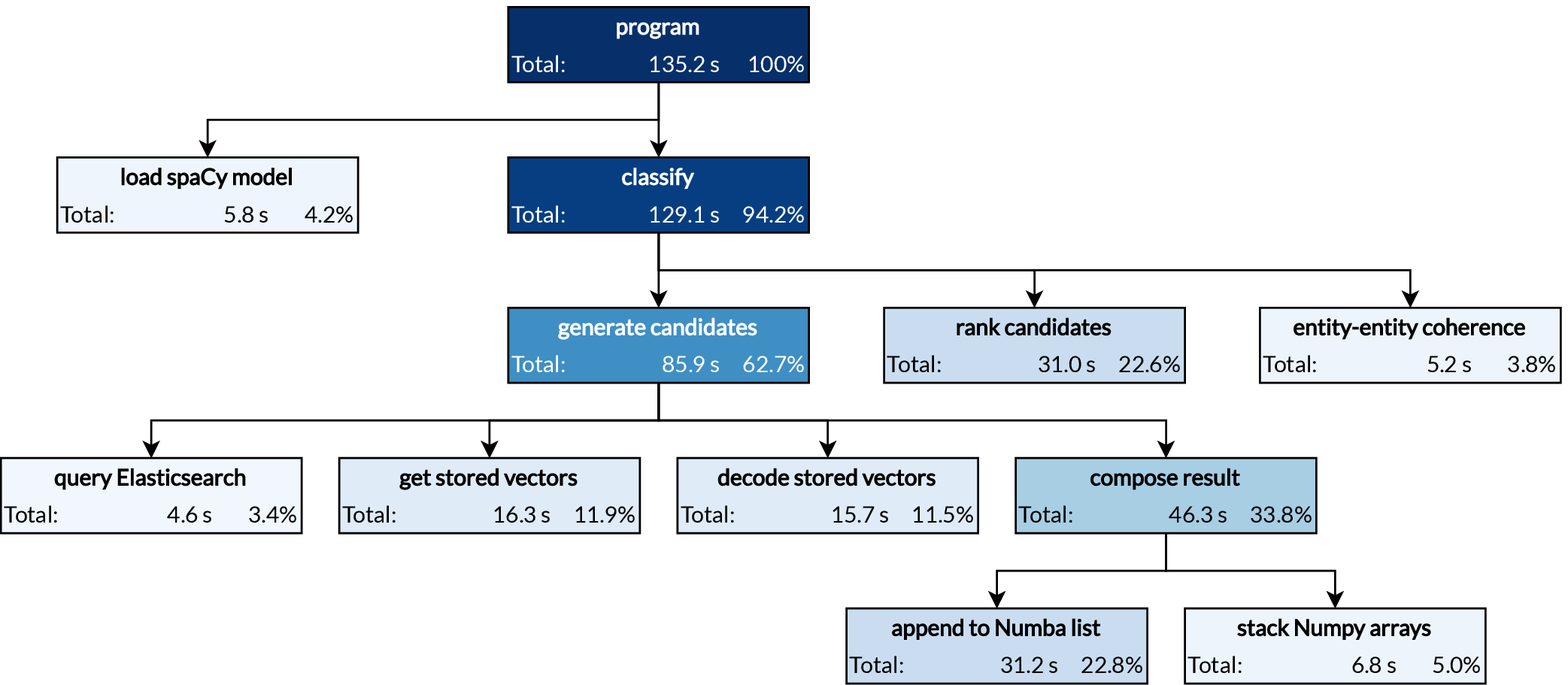}
\centering
\caption{Profiler diagram for classifying 50 EFSA publications in a batch}
\label{fig:profiling}
\end{figure}

To examine possible bottlenecks, the classification of a batch of 50 EFSA publications was profiled using cProfile. The results are presented in~\cref{fig:profiling}, with only the most important calls included.

Over 60\% of the program's execution time is spent generating candidates, which seems counter-intuitive, as this should be simply a matter of querying Elasticsearch. Querying is indeed present here (taking up only 3.4\% of the total time), but for efficiency reasons several other operations are performed as well. LMDB is queried for the stored lexical and semantic vectors (11.9\%). These vectors have to be then decoded with the pickle protocol to Python objects (11.5\%). Finally, the found candidates and their neighbors are rearranged into data structures optimized for the later stages (33.8\%). 

This last step can be especially confusing, raising question why such seemingly simple operation requires so much time. The main reason is that code compiled with Numba, in later stages, requires its input data structures to be Numba-compatible. This necessitates ``boxing'' of Python objects to be performed at some point, which incurs a performance hit. Another option would be not to use Numba, but that would make the program even slower. Another compute-intensive step is candidate ranking, where most of the time is spent in optimized numerical routines.

\section{Conclusions} \label{sec:conclusions}

The proposed method is an adaptable, scalable, and performant solution to the topical publication classification problem, easily handling KBs containing millions of entities. Classification results seem satisfactory, although a detailed evaluation remains to be performed on a labeled dataset. The method is robust, modular base which can be iteratively improved, to achieve better accuracy. In particular, further research into candidate generation and how the algorithm perceives the context of both the abstract and the knowledge base is needed.

Regarding classification speed, substantial improvements could be made by parallelizing the code. However, currently this could not be easily done, due to the Python Global Interpreter Lock (GIL), which prevents the simultaneous execution of multiple Python interpreter threads~\cite{Meier2019GIL}. There is also a significant amount of overhead involved at several steps, necessitated by transitioning from interpreted to native code. The only reasonable solution to these problems would be to drop pure Python as the implementation language altogether in favor of C++ or Cython, which will be investigated in future research.

Additional materials and data related to this work, as well as the most important algorithms mentioned above, were published on GitHub\footnote{\url{https://github.com/Ostrzyciel/food-safety-classif}}.

\renewcommand{\bibsection}{\section*{Bibliography}}
\bibliographystyle{styles/bibtex/spmpsci}
\small{
\bibliography{bib/bibliography}
}

\end{document}